\pdfoutput=1

\documentclass[11pt]{article}

\usepackage{acl}

\usepackage{times}
\usepackage{latexsym}
\usepackage{amsfonts}
\usepackage{amssymb}
\usepackage{amsmath}
\usepackage{amstext}
\usepackage{multirow}
\usepackage{arydshln}
\usepackage[utf8]{inputenc}
\usepackage[arabic,farsi,english]{babel}
\usepackage[pdftex]{graphicx}
\usepackage{pgfplots}
\usepgfplotslibrary{groupplots}

\usepackage[T1]{fontenc}

\usepackage[utf8]{inputenc}

\usepackage{microtype}

\newcommand{\specialcell}[2][c]{%
	\begin{tabular}[#1]{@{}c@{}}#2\end{tabular}}
\newcommand{\specialcellleft}[2][c]{%
	\begin{tabular}[#1]{@{}l@{}}#2\end{tabular}}
	
\makeatletter
\newcommand*{\inlineequation}[2][]{%
    \begingroup
    \refstepcounter{equation}%
    \ifx\\#1\\%
    \else
    \label{#1}%
    \fi
    \relpenalty=10000 %
    \binoppenalty=10000 %
    \ensuremath{%
    	#2%
    }%
    ~\@eqnnum
    \endgroup
}
\makeatother

\title{Persian Natural Language Inference: A Meta-learning approach}

\author{
	Heydar Soudani\Thanks{Equal contribution} \\
	Sharif University of Technology \\
	\texttt{heydars@ce.sharif.edu} \\
  \And
	Mohammad Hassan Mojab\footnotemark[1] \\
	Sharif University of Technology \\
	\texttt{mhmojab@ce.sharif.edu} \\
  \AND
	Hamid Beigy\\
	Sharif University of Technology \\
	\texttt{beigy@sharif.edu} 
}

\begin{document}
\maketitle
\begin{abstract}
Incorporating information from other languages can improve the results of tasks in low-resource languages. A powerful method of building functional natural language processing systems for low-resource languages is to combine multilingual pre-trained representations with cross-lingual transfer learning. In general, however, shared representations are learned separately, either across tasks or across languages. This paper proposes a meta-learning approach for inferring natural language in Persian. Alternately, meta-learning uses different task information (such as QA in Persian) or other language information (such as natural language inference in English). Also, we investigate the role of task augmentation strategy for forming additional high-quality tasks. We evaluate the proposed method using four languages and an auxiliary task. Compared to the baseline approach, the proposed model consistently outperforms it, improving accuracy by roughly six percent. We also examine the effect of finding appropriate initial parameters using zero-shot evaluation and CCA similarity.
\end{abstract}

\section{Introduction}

In natural language processing (NLP), the goal is to improve models for the processing and production of human languages. As part of NLP, several tasks are defined, each covering a different level of natural language understanding. Meanwhile, natural language inference (NLI) is considered an appropriate and rigorous measure of language comprehension. This task requires recognizing the consequences of natural language sentences, which indicates how well it understands the language~\cite{maccartney2009natural}.

NLI aims to determine the inferential relationship between a premise $p$ and a hypothesis $h$. The problem involves a three-class classification in which every pair $(p, h)$ falls into one of three categories: entailment, contradiction, and neutral. If the hypothesis can be inferred from the premise, pair $(p, h)$ will be assigned to the entailment class. For a hypothesis that contradicts the premise, pair $(p, h)$ will be assigned to the contradiction and neutral otherwise~\cite{amirkhani2020farstail}.

The Persian language lacks sufficient linguistic resources when it comes to natural language understanding. The lack of data can be addressed by collecting annotated data, but this process is both time-consuming and expensive~\cite{nooralahzadeh2020zero}.
FarsTail~\cite{amirkhani2020farstail} is currently available for Persian, which is created using the same method as SciTail~\cite{khot2018scitail}. It contains 10,367 samples.
Also, ParsiNLU~\cite{10.1162/tacl_a_00419} is created for high-level tasks in Persian, and for NLI, it consists of 2700 samples. 
As it turns out, this amount of data is too small compared with resource-rich languages (such as English, which has only 550,000 samples in the SNLI~\cite{bowman2015large} dataset).

Researchers have tried to solve the data scarcity problem by using cross-language methods. Recent work on cross-lingual learning has mainly focused on transfer between languages already covered by pre-trained representations~\cite{wu2019beto}.
Nonetheless, these techniques do not readily transfer to low-resource languages in which (1) large monolingual corpora are unavailable for pre-training, and (2) sufficient labeled data is lacking for fine-tuning downstream tasks~\cite{xia2021metaxl}.

The results of experimental studies for Persian using different embedding methods including word2vec~\cite{mikolov2013distributed}, fastText~\cite{bojanowski2017enriching}, ELMo~\cite{peters2018deep}, and BERT~\cite{devlin-etal-2019-bert} and various models, such as, DecompAtt~\cite{parikh2016decomposable}, ESIM~\cite{chen2016enhanced},
HBMP~\cite{talman2019sentence}, and ULMFiT \cite{howard2018universal} is reported in FarsTail~\cite{amirkhani2020farstail}.
Although this cross-lingual information sharing has enabled success in various natural language processing tasks, it raises the question of how we can achieve more effective collaborative learning between languages or even between different tasks.

Recently, meta-learning has shown to be effective for a variety of machine learning tasks, including NLP \cite{koch2015siamese, ravi2016optimization, qian2019domain}.
This paper uses a meta-learning-based method for learning parameters in the joint space of tasks and languages. Auxiliary languages include English, Spanish, French, and German, while QA is the auxiliary task.

Alternatively, an essential prerequisite for the successful application of meta-learning is a task distribution from which a large number of tasks can be sampled to train the meta-learner. However, in NLP, datasets are usually considered as tasks~\cite{nooralahzadeh2020zero, qi2020translation}.
There are two main problems with treating entire datasets as tasks. The first problem is overfitting, in which a meta-learner is overfitted to a small number of training tasks since there is only a small number of supervised datasets for each NLP problem. A second concern is that the heterogeneity of NLP datasets may result in learning episodes that lead to memorization overfitting, where a meta-learner ignores the support set and fails to adapt~\cite{murty2021dreca}. To improve the quality and quantity of tasks, we use the DReCa \cite{murty2021dreca} approach as our data augmentation strategy.
 
In this paper, we employ meta-learning algorithms to enhance the Persian NLI task. Our models are evaluated on the FarsTail dataset. Experimental results show that we push Persian NLI accuracy forward by more than 6\% and zero-shot accuracy by about 4\%, setting a new state-of-the-art result for this task.
In summary, the main contributions of our research are:

\begin{itemize}
	\item We have enabled effective parameter sharing across multiple languages and tasks by providing a meta-learning approach. To the best of the authors' knowledge, this is the first study of the interaction between several languages and tasks at different levels of abstraction to solve a high-level problem in the Persian language. The evaluation results are based on the FarsTail dataset as a reference dataset in the Persian language. The datasets available in the XTREME benchmark~\cite{hu2020xtreme} have also been used for auxiliary languages and tasks.
	
	\item We examine a metadata augmentation strategy named DReCa~\cite{murty2021dreca} that takes as input a set of tasks (entire datasets). We then decompose them to approximate some of the latent reasoning categories underlying these datasets, such as various syntactic constructs within a dataset, or semantic categories such as quantifiers and negation.
	
	\item We also evaluate the trained model in zero-shot mode, which means that the target language (Persian) data is never used during the training process. This test indicates the generality of the model.
\end{itemize}

The rest of the paper is arranged as follows:
Section \ref{sec:related-work} briefly describes related work.
Section \ref{sec:methodology} introduces our method,
and in section \ref{sec:experimental-setup}, we explain the details of the experimental setup.
Section \ref{sec:results} presents practical results.
The results analysis and some justification are described in section \ref{sec:analysis}.
We conclude the paper and summarize future directions in section \ref{sec:conclusion}.

\section{Related Work}\label{sec:related-work}

In this section, we briefly outline related work in three areas. The first area is models based on cross-lingual algorithms. In the second area, we highlight methods based on meta-learning. Finally, we summarize existing data augmentation strategies.

\subsection{Models based on Cross-lingual}
Cross-lingual learning is a method for transferring knowledge from one natural language to another~\cite{pikuliak2021cross}.
Pre-trained models are one of the most widely used examples of cross-lingual learning. Since these models have achieved good results, so \citet{wu2019beto} explored the broader cross-lingual potential of mBERT (multilingual BERT) as a zero-shot language transfer model with five NLP tasks including NLI, covering a total of 39 languages.
Also, \citet{wang2019cross} provides a comprehensive study of the contribution of different components in mBERT to its cross-lingual ability. In addition, it examines the impact of the linguistic properties of the languages, the architecture of the model, and the learning objectives.
\citet{conneau2019cross} proposed two methods for learning cross-lingual language models, one using monolingual data and the other using parallel data and a new cross-lingual language model objective.
\citet{singh2019xlda} introduced a cross-lingual data augmentation method that substitutes part of the input text with its translation in another language.

\citet{huertas2021sml} designed a new architecture called Siamese Inter-Lingual Transformer (SILT), to align multilingual embeddings for NLI efficiently. The paper points out that transformer models are unable to generalize to other domains and have problems with multilingual and inter-linguistic scenarios. A new network has been developed to overcome these weaknesses by combining three parts: a multilingual transformer as pre-trained embedding, an alignment matrix to compute the similarity between two sentences, and a multi-head self-attention block to interpret input strings.


Despite the advances that Cross-lingual methods have made, building NLP systems in these settings is challenging for several reasons. First, the target language does not contain sufficient annotated data for effective fine-tuning. Secondly, pre-trained multilingual representations are not directly transferable due to language disparities~\cite{xia2021metaxl}. 
In contrast to these methods, we consider setting up training models simultaneously on multiple languages and tasks.

\subsection{Meta-learning}
\label{sec:related-work_metalearn}

Meta-learning addresses the problem of learning to learn. By examining many learning problems, a meta-learner learns a model~\cite{liu2020few}. Specifically, the meta-learner uses a meta training set 
$
\mathbb{M} \mathbb{S}=\left\{\left(\mathbb{S}_{i}^{s}, \mathbb{T}_{i}^{s}\right)\right\}_{i=1}^{{N}^{s}},
$ where 
$
\left(\mathbb{S}_{i}^{s}, \mathbb{T}_{i}^{s}\right)
$
are the training (support) and test (query) set of the ${i}^{th}$ learning problem and ${N}^{s}$ is the number of learning problems used for training; and a meta test set 
$
\mathbb{M} \mathbb{T}=\left\{\left(\mathbb{S}_{i}^{t}, \mathbb{T}_{i}^{t}\right)\right\}_{i=1}^{{N}^{t}},
$
 where 
$
\left(\mathbb{S}_{i}^{t}, \mathbb{T}_{i}^{t}\right)
$
are the support and query set of the ${i}^{th}$ test learning problem, while ${N}^{t}$ is the number of learning problems used for the test. Given $\mathbb{M} \mathbb{S}$, the meta-learner learns how to map a pair 
$
\left(\mathbb{S}, \mathbb{T}\right)\
$
into an algorithm that leverages $\mathbb{S}$ to optimally solve $\mathbb{T}$.

Due to the lack of well-defined task distribution, meta-learning has not yet succeeded in NLP, leading to attempts that treat datasets as tasks. An ad hoc task distribution causes problems with quantity and quality. \citet{murty2021dreca} provide a way to break down heterogeneous tasks such as datasets into a set of appropriate subtasks. With this method, data is transferred to the feature space using a pre-trained model. They use k-means to decompose data into $k$ clusters and create tasks by combining these clusters.

Recently, however, the combination of cross-lingual techniques in meta-learning frameworks has also been extensively studied.
To train a model for low-resource languages on NLI and QA tasks, \citet{nooralahzadeh2020zero} uses the MAML algorithm and auxiliary languages.
\citet{van-der-heijden-etal-2021-multilingual} study the text documents classification problem in monolingual and multilingual modes, using different algorithms such as, 
Prototypical Networks \cite{snell2017prototypical}, MAML \cite{finn2017model}, Reptile \cite{nichol2018first}, and ProtoMAML \cite{triantafillou2019meta}.
Also, \citet{tarunesh2021meta} examine the interaction between different languages and tasks to learn an appropriate common feature space.

Additionally, transfer-learning can be helpful for low-resource languages. \citet{xia2021metaxl} introduce a meta-learning-based framework called MetaXL for extremely low-resource languages.
MetaXL learns an intelligent representational conversion from several auxiliary languages to the target language, bringing the feature space of these languages closer together for more efficient conversion. The main idea is to use a Representation Transformation network between the main model layers which are trained only with target language.

To the best of our knowledge, this paper is the first attempt to study meta-learning for solving the NLI problem in the Persian language. Also, we are pioneers in using task-language pairs as meta-learning tasks in the Persian language. 

\subsection{Task Augmentation}

Machine learning algorithms usually assume that the train and test data have the same distribution. In contrast, the meta-learning framework treats tasks as training examples and trains a model to adapt to all of them. Meta-learning also assumes that the training and new tasks are drawn from the same distribution of tasks $p(\tau)$. In NLP, datasets are typically treated as tasks, and meta-learners are then overfitting their adaptation mechanisms. NLP datasets are highly heterogeneous, which causes many learning episodes to have the poor transfer between their support and query sets, which dissuades meta-learners from adapting~\cite{murty2021dreca}.

To deal with overfitting challenges, \citet{yin2019meta} propose a meta-regularizer to mitigate memorization overfitting, but don’t study learner overfitting.
\citet{rajendran2020meta} study task augmentation for mitigating meta-learners overfitting in the context of few-shot label adaptation.
SMLMT method \cite{bansal2020self} creates new self-supervised tasks that improve meta-overfitting, but this does not directly address the dataset-as-tasks problem.
In contrast, the DReCa method \cite{murty2021dreca} addresses the dataset-as-tasks problem and focuses on using clustering as a way to subdivide and fix tasks that already exist. In this paper, we use DReCa as a task augmentation strategy for our method since it mitigates meta-overfitting without any additional unlabeled data.

\renewcommand{\arraystretch}{1.2}
\begin{table}[!ht]
	\centering
	\begin{footnotesize}
	\begin{tabular}{c|cc}
		\hline 
		 & \textbf{NLI} & \textbf{QA} \\ 
		\hline
		\textbf{FA} & FarsTail (10.3K) & PersianQA (9K) \\
		\textbf{EN} & XNLI (392k)      & --- \\
		\textbf{ES} & tr. XNLI (392k)  & --- \\
		\textbf{DE} & tr. XNLI (392k)  & --- \\
		\textbf{FR} & tr. XNLI (392k)  & --- \\
		\hline
	\end{tabular}
	\end{footnotesize}
	\caption{Overview of datasets from a variety of sources. For the NLI task, we use the XNLI dataset for English, and its translated versions (tr.) for Spanish(ES), German(DE), and French(FR) provided in XTREME. For each dataset, the number of training instances is also mentioned.}
	\label{table:datasets}
\end{table}

\section{The Proposed Methodology}
\label{sec:methodology}

In our setting, firstly, we prepare a set of task-language pairs to provide meta-learning tasks. Afterward, in each episode, we sample some tasks and feed them to the meta-learner. In the rest of this section, we describe the proposed task sampling strategy, the proposed meta-learning algorithm, and the proposed task augmentation strategy.

\subsection{The Proposed Task Sampling Strategy}
\label{sect:met1}

In meta-learning, task selection has a profound impact on model performance.  For this reason, we create a queue of tasks first.  We can create this queue using different scenarios such as selecting languages for a target task~\cite{gu2018meta}, selecting tasks for a target language~\cite{dou2019investigating}, and picking from various auxiliary languages and auxiliary tasks. In the meta-training section, we sample some tasks from the queue.
Formally, the queue's tasks are represented by ${\mathcal{D}}$.  We need to sample tasks from ${\mathcal{M}}$, which is a Multinomial distribution over $P_{\mathcal{D}}(i)$s. Thus, we investigate temperature-based heuristic sampling \cite{arivazhagan2019massively}, which defines the probability of any dataset as a function of its size as,
\begin{equation}
P_{\mathcal{D}}(i)=q_{i}^{1 / \tau} /\left(\sum_{k=1}^{n} q_{k}^{1 / \tau}\right)
\end{equation}
where $P_{\mathcal{D}}(i)$ is the probability of sampling the $i$th task, $q_{i}$ is the size of $i$th task, and $\tau$ is the temperature parameter.
With $\tau = 1$, tasks are randomly sampled proportionately to their dataset sizes, and with $\tau \rightarrow \infty$, they follow a uniform distribution.

\subsection{The Proposed Meta-learning Algorithms}
\label{sect:met2}

Meta-learning is the process of building a model that can solve a new task with only a few labeled examples by training on a variety of tasks with rich annotations. The key idea is to train the model’s initial parameters such that the model has maximal performance on a new task after the parameters have been updated through zero or a couple of gradient steps \cite{yin2020meta}. 
MAML (model-agnostic meta-learning) \cite{finn2017model} is one of the most significant algorithms.
We describe one episode of the MAML algorithm in Appendix~\ref{app:maml}.
MAML is quite difficult to train, since there are two levels of training. Therefore, we use the following two optimization-based and metric-based meta-learning algorithms in this work. 

\textbf{Reptile} \cite{nichol2018first} is a first-order optimization-based algorithm that moves weights toward a manifold of the weighted averages of task-specific parameters $\theta_{i}^{(m)}$.
It samples training tasks from
$p(\mathcal{T}): \tau_{1}, \cdots, \tau_{i}, \cdots, \tau_{n}$.
For each training task, it generates an episode that just contains the support set data.
For training task ${\tau}_{i}$, let’s assume the original parameters $\theta$ have gone through $m$ steps of updating and become ${\theta}_{i}^{(m)}$
(i.e., $\inlineequation[eq:meta-step]{\theta_{i}^{(m)}=\operatorname{AdamW}\left(L_{\tau_{i}}, \theta, m\right)})$, then Reptile updates $\theta$ as follows \cite{yin2020meta}:
\begin{equation}
\theta \leftarrow \theta+\beta \frac{1}{\left|\left\{\mathcal{T}\right\}\right|} \sum_{\tau_{i} \sim \mathcal{M}}\left(\theta_{i}^{(m)}-\theta\right)
\end{equation}

\textbf{Prototypical Networks} \cite{snell2017prototypical} is a metric-based meta-learning algorithm. Prototypical networks learn a metric space in which classification can be performed by computing distances to prototype representations of each class.
In general, they are composed of an embedding network $f_{\theta}$ and a distance function $d\left(x_{1}, x_{2}\right)$. Using the following equation, the embedding network encodes the support set samples $S_{c}$ and computes prototypes $\mu_{c}$ per class based on the mean sample encodings for that class.
\begin{equation}
    \mu_{c}=\frac{1}{\left|S_{c}\right|} \sum_{\left(x_{i}, y_{i}\right) \in S_{c}} f_{\theta}\left(x_{i}\right).
\end{equation}

A Prototypical network classifies a new sample according to the following rule.
\begin{equation}
    p(y=c \mid x)=\frac{\exp \left(-d\left(f_{\theta}(x), \mu_{c}\right)\right)}{\sum_{c^{\prime} \in C} \exp \left(-d\left(f_{\theta}(x), \mu_{c^{\prime}}\right)\right)}
\end{equation}

Thus, we define the \emph{distance-based cross entropy} (DCE) loss as follows:
\begin{equation}
\operatorname{Loss}(DCE)=-\log P\left(y=c \mid x\right)
\end{equation}

To ensure that the feature space is robust to noise, we also use the Cross Entropy (CE) loss (more details can be found in Appendix~\ref{app:pt_loss}).

\subsection{The Task Augmentation Strategy}
\label{sect:met3}

First, we use dataset-as-tasks strategy that is the most common method for selecting tasks for meta-learning in NLP applications.
Next, we employ DReCa to form additional high quality tasks. The goal of DReCa is to take a heterogeneous task (such as a dataset) and produce a decomposed set of tasks.
Given a training task $T_{i}^{tr}$, DReCa first groups examples by their labels, and then embeds examples within each group with an embedding function EMBED(.). Concretely, for each $N$-way classification task $T_{i}^{tr}$, it forms groups
$
g_{l}^{i} = \left\{\left(\operatorname{EMBED}\left(x_{i}\right), y_{i}\right) \mid y_{i}=l\right\}
$.
Then, it proceeds to refine each label group into $K$ clusters via k-means clustering to break down $T_{i}^{tr}$ into groups
$
\left\{C^{j}\left(g_{l}^{i}\right)\right\}_{j=1}^{K} \text { for } l=1,2, \ldots, N.
$ 
These cluster groups can be used to produce $K^{N}$ potential DReCa tasks. Each task is obtained by choosing one of $K$ clusters for each of the $N$ label groups, and taking their union.

\section{Experimental Setup}
\label{sec:experimental-setup}

\subsection{Datasets}

We use FarsTail \cite{amirkhani2020farstail} for the target dataset. FarsTail is the only large-scale Persian corpus for the NLI task, with 10,367 samples. 
The samples are generated from 3,539 multiple-choice questions with the least amount of annotators' interventions or selected from natural sentences that already exist independently in the wild, similarly to the SciTail dataset \cite{khot2018scitail}.

We also use XTREME~\cite{hu2020xtreme} as an auxiliary dataset. XTREME is a multilingual multi-task benchmark consisting of classification, structured prediction, QA, and retrieval tasks.
We use this benchmark to prepare NLI data for auxiliary languages.
Note that, large-scale datasets for NLI were only available in English. However, the authors of XTREME developed a custom-built translation system to get translated datasets for NLI.
Furthermore, we consider the QA as an auxiliary task. Therefore, we use PersianQA \cite{PersianQA} which is a Persian reading comprehension dataset for QA, containing over 9000 entries.
Table~\ref{table:datasets} summarizes the employed dataset specifications.

\subsection{Baselines}

On the FarsTail dataset, \citet{amirkhani2020farstail} present results of various traditional and deep learning-based methods. According to the results of this paper, the highest test accuracy is obtained by using a translation-based approach, i.e., \textit{Translate-Source} with fastText embeddings. In \textit{Translate-Source}, the Persian-translated MultiNLI training set is combined with FarsTail training data for training an ESIM model.
Furthermore, FarsTail's authors reported mBERT fine-tuning results in FarsTail webpage\footnote{\url{https://github.com/dml-qom/FarsTail}}.
Therefore, we use these results as baselines.

\subsection{Implementation Details}

In this study, we aim to compare the effects of meta-learning algorithms on classification accuracy with those of fine-tuning and non-episodic algorithms.
To make a fair comparison, we first fine-tune our pre-trained models using training data of the auxiliary task in a non-episodic approach. Afterward, we fine-tune the obtained model using the training data of the target task. In this approach, we use mBERT \cite{devlin-etal-2019-bert}, and XLM-R \cite{conneau2020unsupervised}, which are known as the state-of-the-art multilingual pre-trained models, and ParsBERT \cite{farahani2020parsbert} as a monolingual transformer-based model for the Persian language.

In the meta-learning approach, we use the XLM-R model with output layers tailored for each task and train it with Reptile and Prototypical algorithms.
To select the hyperparameters of the Reptile algorithm, we utilize the experiments done in \citet{tarunesh2021meta}. Appendix~\ref{app:hyper_parameters} provides further details.
The Prototypical algorithm is used only in cross-lingual experiments, and we use Euclidean distance as its distance function.
The auxiliary languages are arranged in two scenarios.
In the first scenario, support and query set data are generated from auxiliary languages, while in the second scenario, the query set is drawn from both auxiliary and target languages. Detailed information is provided in Appendix~\ref{app:pt_scenarios}.

Furthermore, we fine-tune the obtained models on Persian training data using the following two methods. The first method is \textbf{non-episodic}, which involves fine-tuning models in batches. The second method is \textbf{episodic}, in which episodes are constructed first, and then the models are fine-tuned according to the algorithm used.

\section{Results}
\label{sec:results}

The meta-learning model is tested on different combinations and configurations of the auxiliary tasks.
The accuracy results of the Reptile algorithm are presented in Table~\ref{table:main_results}. 
In addition to the zero-shot and fine-tuning results, we report the accuracy of another scenario. In this scenario, training data of the target language is placed in the meta-training stage along with other auxiliary tasks and cooperate in a training process. Consequently, this scenario does not involve fine-tuning phase.
The results of the mentioned scenario are shown in the last column of Table~\ref{table:main_results}.

\renewcommand{\arraystretch}{1.05}
\begin{table*}[!ht]
	\centering
    \begin{footnotesize}
    	\begin{tabular}{c|c|ccccc}
    		\hline
    		Row
    		& Model
    		& Shot
    		& Aux. Tasks
    		& Zero-shot
    		& Non-episodic f.t.
    		& Add NLI-fa in m.t. \\
    		\hline
    		\hline
    		\multicolumn{7}{c}{Baselines} \\ 
    		\hline 
    		1 & Translate-Source$^*$
    		 & --- & --- & --- & 78.13 & --- \\
    		2 & mBERT$^*$
    		& --- & --- & --- & 83.38 & --- \\
    		\hline
    		\multicolumn{7}{c}{Non-episodic approach} \\
    		\hline
    		3 & ParsBERT
    		& --- & --- & --- & 74.64 & --- \\
    		\cline{3-7}
    		4 & \multirow{3}{*}{mBERT}
    		& --- & ---                                     & --- & 81.95 & --- \\
    		5 & & --- & NLI-en 				                    & 56.53 & 81.38 & --- \\
    		6 & & --- & \specialcell{{NLI-(en, es, de, fr)}} & 67.88 & 82.34 & --- \\
    		\cline{3-7}
    		7 & \multirow{3}{*}{XLM-R} & --- & --- & --- & 81.97 & --- \\
    		8 & & --- & NLI-en & {\bf 69.49} & {\bf 86.55} & --- \\
    		9 & & --- & \specialcell{{NLI-(en, es, de, fr)}} & 69.09 & 84.69 & --- \\
    		\hline
    		
    		\multicolumn{7}{c}{Meta-learning approach}\\
    		\hline
    		10 & \multirow{16}{*}{XLM-R} &1& \multirow{4}{*}{NLI-en}& 64.19 & 84.31 & 83.37 \\
    		11 & &4&  						    & 70.96 & 87.17 & 86.00 \\
    		12 & &8&  					        & 70.70 & 87.11 & 86.65 \\
    		13 & &16&  					        & 71.03 & 87.43 & 86.52 \\
    		\cline{3-7}
    		14 & &1 & \multirow{4}{*}
    		     {\specialcell{{NLI-(en, es, de, fr)}}}   & 65.17 & 85.21       & 83.91 \\
    		15 & & 4 & & {\bf 72.27} & 87.57 & 85.74 \\
    		16 & & 8 & & 71.61 & {\bf 88.35} & {\bf 88.22} \\
    		17 & &16 & & 71.22 & 88.02       & 87.76 \\
    		\cline{3-7}
    		18 & &1& \multirow{4}{*}{QA-fa}      & 34.18 & 81.48 & 81.58 \\
    		19 & &4&  						    & 34.18 & 81.38 & 84.96\\
    		20 & &8&  					        & 33.79 & 82.14 & 83.59\\
    		21 & &16&  					        & 34.18 & 82.23 & 84.70\\
    		\cline{3-7}
    		22 & &1 & \multirow{4}{*}
    		     {\specialcell{{NLI-en, QA-fa}}} & 46.42 & 83.53 & 85.16\\
    		23 & &4&  						 	        & 66.02 & 86.52 & 86.26\\
    		24 & &8&  					                & 64.26 & 86.98 & 86.46\\
    		25 & &16&  					                & 46.88 & 86.52 & 86.13\\
    		\hline
    	\end{tabular}
	\end{footnotesize}	
	\caption{Average test accuracy of the Reptile algorithm with baselines and non-episodic approach results on the Persian NLI task. The first accuracy column shows results before fine-tuning on the Persian NLI train-set (called zero-shot). In the second accuracy column, we provided results after fine-tuning (f.t.) on the Persian NLI train-sets. The last accuracy column reports results of using the Persian NLI train-set in the meta-training phase (m.t.). The data with $^*$ comes from FarsTail's paper and webpage.}
	\label{table:main_results}
\end{table*}

Table~\ref{table:main_results_pt} shows the accuracy scores using the Prototypical Network.
In the first section of this table, we generate both support and query sets from Persian language data, without using auxiliary tasks.
In the second section of this table, the results of the first multi-lingual scenario (where both the support and query sets are generated from auxiliary languages data) are reported in rows 5 to 12.
In rows 13 to 16, we show the results of the second multi-lingual scenario (where the support set is drawn from auxiliary language data and the query set is drawn from both auxiliary and Persian language data).
Lastly, we added the DReCa strategy and presented the results in rows 17 to 20.

\begin{table*}[!ht]
	\centering
	\begin{footnotesize}
    	\begin{tabular}{c|c|cccccc}
    	\hline
    	Row
    	& Model
    	& Shot
    	& Support
    	& Query
    	& Zero-shot
    	& Non-episodic f.t.
    	& Episodic f.t. \\
    	\hline
    	\hline 
    	\multicolumn{8}{c}{Without auxiliary tasks}\\
    	\hline 
    	1 & \multirow{4}{*}{XLM-R} &1& \multirow{4}{*}{NLI-fa} & \multirow{4}{*}{NLI-fa} & --- & 70.38 & 79.30\\
    	2 & & 4 & &	& --- & 81.97 & 85.22\\
    	3 & & 8 & & & --- & 83.98 & 84.64\\
    	4 & & 16& & & --- & 85.29 & 85.74\\
    	\hline
    	\multicolumn{8}{c}{With auxiliary tasks}\\
    	\hline
    	5 & \multirow{12}{*}{XLM-R} & 1 & \multirow{4}{*}{NLI-en} & \multirow{4}{*}{NLI-en} & 68.10 & 84.83 & 86.07\\
    	6 & & 4 & & & 70.57 & 86.72 & 87.50 \\
    	7 & & 8 & &	& 70.77 & 86.72 & 87.37 \\
    	8 & & 16& &	& {\bf73.18} & 87.76 & 88.54 \\
    	\cline{3-8}
    	9 & &1& \multirow{4}{*}{\specialcell{{NLI-(en, es, de, fr)}}}
    		& \multirow{4}{*}{\specialcell{{NLI-(en, es, de, fr)}}}
    		                     			& 69.15 & 85.01 & 85.97\\
    	10 & &4& & & 70.25 & 86.78 & 87.63 \\
    	11 & &8& & & 71.09 & {\bf88.48} & {\bf89.39}\\
    	12 & &16 & & & 72.20 & 88.15 & 88.28 \\
    	\cline{3-8}
    	13 & &1& \multirow{4}{*}{\specialcell{{NLI-(en, es, de, fr)}}}
    	& \multirow{4}{*}{\specialcell{{NLI-(en, es, de, fr, fa)}}}
    	                     				& --- & 84.15 & 85.12\\
    	14 & &4&  				&		    	& --- & 86.33 & 86.78\\
    	15 & &8&  				&	            & --- & 86.33 & 86.46\\
    	16 & &16&  				&				& --- & 86.78 & 87.24\\
    	\hline
    	17 & \multirow{4}{*}{\specialcell{{XLM-R+}\\{DReCa}}} 
    	& 8 & \multirow{2}{*}{NLI-en} & \multirow{2}{*}{NLI-en} & 70.44 & 87.96 & 88.87 \\
    	18 & & 16 &  & & 71.94 & 87.24 & 88.74 \\
    	\cline{3-8}
    	19 & & 8 & \multirow{2}{*}{\specialcell{{NLI-(en, es, de, fr)}}} & \multirow{2}{*}{\specialcell{{NLI-(en, es, de, fr)}}} & 71.16 & 87.74 & 88.48 \\
    	20 & & 16 & & & 71.61 & 87.30 & 88.22 \\
    	\hline
    	\end{tabular}
	\end{footnotesize}
	\caption{Average test accuracy on the Persian NLI task using Prototypical algorithm with and without auxiliary tasks. The last accuracy column reports results after episodic fine-tuning (f.t.) on the Persian NLI train-set.}
	\label{table:main_results_pt}
\end{table*}

Additionally, we conducted zero-shot evaluations of both algorithms.
Zero-shot results are presented in the first accuracy column of Tables~\ref{table:main_results} and \ref{table:main_results_pt}.
The confusion matrices of the best-performing models for both Reptile and Prototypical algorithms are also depicted in Appendix~\ref{app:confusion_matrices}.

\section{Discussion and Analysis}
\label{sec:analysis}

Table~\ref{table:main_results} shows that the multi-lingual models are always better than the multi-task models, due to the fact that tasks like NLI (which require deeper semantic representations) are more likely to benefit from combining data from different languages.
We found that our meta-learned models perform better than baselines and non-episodic models. The reason is that the goal of standard meta-learning is to find a model that generalizes well to a new target task.
In addition, we compared two different meta-learning algorithms to evaluate their superiority in this paper. From Tables \ref{table:main_results} and \ref{table:main_results_pt}, we can see that Prototypical performed better than Reptile. It is because Prototypical networks use class representations instead of example representations. Therefore, it finds a suitable representation for each class during the meta-train stage.

As part of another experiment, we combined data from the target language with data from other auxiliary tasks for meta-training. Based on the results of these experiments (last column of Table~\ref{table:main_results} for Reptile and rows 13 to 16 of Table~\ref{table:main_results_pt} for Prototypical), the model's accuracy has decreased. This is due to the fact that target language data is small when compared with auxiliary language data.
So, unbalanced training data confuses the training process and decreases the model's accuracy.
In any case, the cooperation of the target language during the training process is a great idea for future work.

As indicated in the last two columns of Table~\ref{table:main_results_pt}, episodic fine-tuning is significantly superior to non-episodic fine-tuning. It demonstrates that episodic training is effective even on single language data and creates a generality in the level of training and test data.

\begin{figure*}[!ht]
    \centering
    \includegraphics[width=\textwidth]{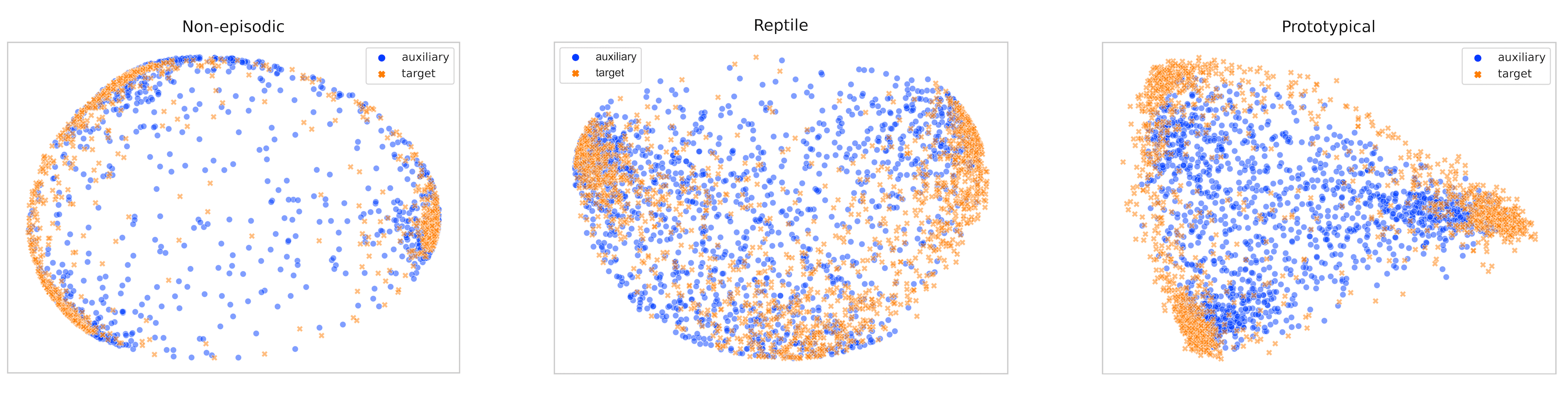}
    \vspace*{-5mm}
    \caption{PCA visualization of non-episodic, Reptile, and Prototypical models to examine the closeness of the auxiliary and target languages feature spaces.}
    \label{fig:pca}
\end{figure*}

We examined the proximity between the feature spaces of the auxiliary languages and the target language quantitatively and qualitatively.
At first, we collect representations of the auxiliary and target languages from non-episodic, Reptile, and Prototypical models.
In Fig.~\ref{fig:pca}, we present 2-component PCA visualization for comparison.
We also evaluated the models using a distance metric commonly used in vision and NLP tasks~\cite {huttenlocher1993comparing, dubuisson1994modified, patra2019bilingual, xia2021metaxl}.
Informally, the Hausdorff distance measures the distance between data representations of auxiliary languages and the target language.
Given a set of representations of the auxiliary language $\mathcal{S}=\left\{s_{1}, s_{2}, \ldots, s_{m}\right\}$ and a set of representations of the target language $\mathcal{T}=\left\{t_{1}, t_{2}, \ldots, t_{m}\right\}$ we compute the Hausdorff distance as follows:
\begin{equation}
\max \left\{\max _{s \in \mathcal{S}} \min _{t \in \mathcal{T}} d(s, t), \max _{t \in \mathcal{T}} \min _{s \in \mathcal{S}} d(s, t)\right\}
\end{equation}
where cosine distance is used as the inner distance, i.e.,
\begin{equation}
d(s, t) \triangleq 1-\cos (s, t)
\end{equation}

Compared to the non-episodic method, we observe a drastic drop of Hausdorff distance from 0.18 to 0.05 for Prototypical and also, we see a minor decline of Hausdorff distance from 0.18 to 0.13 for Reptile.
Both qualitative visualization and quantitative metrics confirm that meta-learning approaches bring the distributions of auxiliary and target language data closer together, thus increasing the accuracy on the target language.

The advantage of meta-learning methods is that they obtain the appropriate initial parameters for the target language, as mentioned. The zero-shot test is used as a criterion to evaluate this point, and it shows that meta-learning-based models are more accurate than other methods. The generality of the initial parameters can also be assessed via canonical correlation analysis (CCA) \cite{NIPS2017_7188, NIPS2018_7815}. Using this criterion, we compare the output of each layer before and after fine-tuning, and the results are presented in Fig.~\ref{fig:cca_result}. The meta-learning models have a higher CCA similarity, which indicates the model obtained more general parameters before fine-tuning.

\pgfplotsset{compat=1.3}
\begin{figure}[!ht]
    \centering
    \resizebox{0.485\textwidth}{!}{
        \begin{tikzpicture}
            \tikzstyle{every node}=[font=\tiny]
            \pgfmathsetlengthmacro\MajorTickLength{
              \pgfkeysvalueof{/pgfplots/major tick length} * 0.4
            }
            \begin{axis}[
                xlabel={layer},
                ylabel={CCA similarity},
                xmin=-0.5, xmax=11.5,
                ymin=0.6, ymax=1.01,
                xtick={0,1,2,3,4,5,6,7,8,9,10,11},
                ytick={0.6,0.7,0.8,0.9,1.0},
                xtick pos=left,
                ytick pos=left,
                xtick align=outside,
                ytick align=outside,
                major tick length=\MajorTickLength,
                xlabel near ticks,
                ylabel near ticks,
                legend pos=south west,
                legend cell align=left,
                legend image post style={scale=0.8},
                legend style={row sep=-3.3pt, at={(0.01,0.01)}, inner ysep=0pt, inner xsep=2pt},
            ]
            
            \addplot[
                color=green,
                mark=diamond*,
                mark options={scale=0.80}
                ]
                coordinates {
                    (0,0.9978260831197882)
                    (1,0.9900747836588831)
                    (2,0.9838725951201711)
                    (3,0.9757968202838009)
                    (4,0.9654524428851016)
                    (5,0.955005947733535)
                    (6,0.9403452068442736)
                    (7,0.9297261678086207)
                    (8,0.8873356301871023)
                    (9,0.8458321812284316)
                    (10,0.7662655311561677)
                    (11,0.7464979584114264)
                };
            
            \addplot[
                color=violet,
                mark=x,
                mark options={scale=0.80}
                ]
                coordinates {
                    (0,0.9974956455351037)
                    (1,0.9869140100273198)
                    (2,0.9767378996450381)
                    (3,0.9679597455659769)
                    (4,0.9570889326947166)
                    (5,0.9489488906495008)
                    (6,0.9382154343238454)
                    (7,0.9182930637022689)
                    (8,0.8726146703446743)
                    (9,0.8389094726521034)
                    (10,0.7755252623768181)
                    (11,0.7462087908288669)
                };
                
            \addplot[
                color=brown,
                mark=triangle*,
                mark options={scale=0.80}
                ]
                coordinates {
                    (0,0.996192407371848)
                    (1, 0.982399756320465)
                    (2, 0.9648715704655665)
                    (3, 0.9543579150980852)
                    (4, 0.9343621200958866)
                    (5, 0.9246303679369109)
                    (6, 0.9057134277192055)
                    (7, 0.88393163073651)
                    (8, 0.8294765909907253)
                    (9, 0.7752323400608034)
                    (10, 0.7175239119393696)
                    (11, 0.7099931882299408)
                };
            
            \addplot[
                color=blue,
                mark=|,
                mark options={scale=0.80}
                ]
                coordinates {
                    (0,0.9963366852226092)
                    (1, 0.980412584042606)
                    (2, 0.9651295127503717)
                    (3, 0.9526558179457126)
                    (4, 0.9390529307006842)
                    (5, 0.9295803032707505)
                    (6, 0.9128913452512503)
                    (7, 0.8938212243908265)
                    (8, 0.8000321090801356)
                    (9, 0.7658295450121825)
                    (10, 0.7291885679741771)
                    (11, 0.7054413174669053)
                };
                
            \addplot[
                color=purple,
                mark=square*,
                mark options={scale=0.80}
                ]
                coordinates {
                    (0,0.9968445518521888)
                    (1,0.9827014147553522)
                    (2,0.9614060765438898)
                    (3,0.9387345081820319)
                    (4,0.9205085691941529)
                    (5,0.9006617367880692)
                    (6,0.8818581914572393)
                    (7,0.8637575997704574)
                    (8,0.7962959336439903)
                    (9,0.7494117948594855)
                    (10,0.7058843318498922)
                    (11,0.675215592789189)
                };
                
            \addplot[
                color=darkgray,
                mark=star,
                mark options={scale=0.80}
                ]
                coordinates {
                    (0,0.9960603187335559)
                    (1,0.9801339315207048)
                    (2,0.9569743527638649)
                    (3,0.9297731281640361)
                    (4,0.9058912969910408)
                    (5,0.8924045386082659)
                    (6,0.8708411667158665)
                    (7,0.8580053862413699)
                    (8,0.7955811240022209)
                    (9,0.740074158712195)
                    (10,0.6969297142284184)
                    (11,0.6810987694452476)
                };
                
            \addplot[
                color=lightgray,
                mark=*,
                mark options={scale=0.80}
                ]
                coordinates {
                    (0,0.9535018842326805)
                    (1,0.9336592956526056)
                    (2,0.8974597892120914)
                    (3,0.8772908926673537)
                    (4,0.8245836852939218)
                    (5,0.7766242644800921)
                    (6,0.7644845555573542)
                    (7,0.746254565470776)
                    (8,0.7259084485389197)
                    (9,0.7078838277221133)
                    (10,0.6950564479433542)
                    (11,0.6737965381909125)
                };
                
            \legend{
                {} {Reptile nli-(en,fr,es,de)},
                {} {Reptile nli-(en)},
                {} {Prototypical nli-(en,fr,es,de)},
                {} {Prototypical nli-(en)},
                {} {Non-episodic XLM-R nli-(en)},
                {} {Non-episodic XLM-R nli-(en,fr,es,de)},
                {} {Non-episodic mBERT nli-(en,fr,es,de)}
            }
                
            \end{axis}
        \end{tikzpicture}
    }
    \vspace*{-5mm}
    \caption{CCA similarity for each transformer layer. We calculate the similarity before and after fine-tuning on the FarsTail training data.}
    \label{fig:cca_result}
\end{figure}
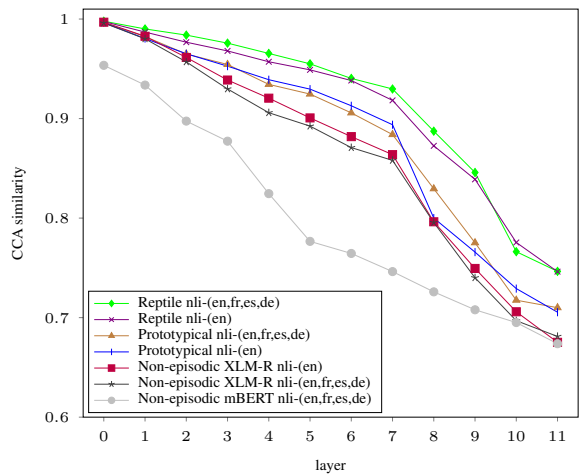

In the next experiment, we apply the DReCa strategy and train the model with the Prototypical algorithm. According to Table~\ref{table:main_results_pt}, some results have improved, while others have remained the same. It illustrates that task augmentation in meta-algorithms affects the model's accuracy. However, defining the appropriate task augmentation strategy still needs research.

\section{Conclusion}
\label{sec:conclusion}

We present effective use of meta-learning to benefit from other tasks or languages. We advantageously leverage this approach to improve NLI in Persian as a low-resource language.
We found that our meta-learning model outperformed competitive baseline models.
In response to the concept of treating entire datasets as tasks, we use DReCa as a general-purpose task augmenting approach.
Finally, zero-shot evaluations illustrate the generality of the results obtained by meta-learning.
This work will be extended to other cross-lingual NLP tasks in Persian in the future. Furthermore, we would like to use a self-supervised approach to provide a useful starting point for parameters.

\bibliography{custom}

\appendix

\appendix
\renewcommand\thesubsection{A.\arabic{subsection}}

\section{Appendix} 

\subsection{MAML Description}\label{app:maml}
MAML is one of the most popular meta-learning algorithms and it has proven its effectiveness in various fields (e.g., computer vision).
MAML is able to find good initialization parameter values and adapt to new tasks quickly.
This algorithm can be performed in one episode by following these steps:
\begin{itemize}
	\item Make a copy of the model with its initial parameters $\theta$.
	\item Use the training set $D_i^{train}$ to train the model as
    	\begin{equation}
    	\hat{\theta}=\theta-\alpha \nabla_{\theta} \mathcal{L}_{i}\left(\theta, \mathcal{D}_{i}^{{train }}\right)
    	\end{equation}
	\item Apply the model with the updated parameters $\hat{\theta}$ to the validation set $D_i^{val}$.
	\item Use the loss on the validation set to update the initial parameters $\theta$
		\begin{equation}
		\theta=\theta-\beta \nabla \theta \sum_{i} \mathcal{L}_{i}\left(\hat{\theta}, \mathcal{D}_{i}^{val}\right)
		\end{equation}
	
\end{itemize}

\subsection{Hyperparameters}
\label{app:hyper_parameters}

Models are implemented using the PyTorch\footnote{\url{https://pytorch.org/}} framework. 
ParsBERT, mBERT and XLM-R implementations are taken from the HuggingFace library \footnote{\url{https://huggingface.co/}}.

In our experiments, we used the AdamW optimizer~\cite{loshchilov2018fixing} with learning rate 1e-5 to perform the inner loop of the Reptile algorithm \eqref{eq:meta-step}, which is known as meta-step. The hyperparameters for the Reptile algorithm are listed in Table~\ref{table:reptile_hp}.

\begin{table}[!ht]
	\centering
	\begin{tabular}{lr}
    \hline
	Hyperparameter & Value \\
    \hline
    epochs                              & 2     \\
    number of iterations                & 20000 \\
    sequence length (for NLI)           & 128   \\
    sequence length (for QA)            & 384   \\
    dropout                             & 0.1   \\
    optimizer                           & AdamW \\
    learning rate                       & 1e-5  \\
    update steps (m)                    & 3     \\
    number of class per episode (way)   & 2     \\
    queue length                        & 4     \\
    temperature parameter ($\tau$)      & 1     \\
    \hline
    \end{tabular}
	\caption{Hyperparameters for the Reptile algorithm}
	\label{table:reptile_hp}
\end{table}

The hyperparameters for the Prototypical algorithm are also shown in Table~\ref{table:pt_hp}. Some parameters are calculated based on a grid search, such as Distance Cross-Entropy (DCE) and Cross-Entropy (CE) coefficients, and others are chosen similar to the Reptile algorithm.

\begin{table}[!ht]
	\centering
	\begin{tabular}{lr}
    \hline
	Hyperparameter & Value \\
    \hline
    epochs                              & 2     \\
    number of iterations                & 20000 \\
    sequence length (for NLI)           & 128   \\
    dropout                             & 0.1   \\
    optimizer                           & AdamW \\
    learning rate                       & 1e-5  \\
    number of class per episode (way)   & 3     \\
    DCE coefficient ($\lambda_{1}$)     & 1.0   \\
    CE coefficient ($\lambda_{2}$)      & 1.0   \\
    \hline
    \end{tabular}
	\caption{Hyperparameters for the Prototypical algorithm}
	\label{table:pt_hp}
\end{table}

The number of iterations parameter varies according to the value of the shot, and is chosen to ensure that all instances in the dataset appear at least once in each epoch.

\subsection{Prototypical Networks}

\subsubsection{Loss Function}
\label{app:pt_loss}

As we mentioned in section~\ref{sec:related-work_metalearn}, the primary loss function of the Prototypical algorithm is DCE.
Since a prototype consists of distribution information from instances associated with it, the choice of these instances may introduce noise in the learned representation if the neural network is trained only by using the DCE loss. We use CE loss in addition to the DCE loss to make the feature space robust to noise. As a whole, we train the model with a combination of DCE loss and CE loss given by the following equation.
\begin{equation}
\text{Loss}(\text{overall})=\lambda_{1}\operatorname{Loss}(DCE)+\lambda_{2} \operatorname{Loss}(CE)
\end{equation}

\subsubsection{Scenarios}
\label{app:pt_scenarios}

We considered two scenarios for making the episodes.
In the first scenario, the model is trained only on auxiliary languages, then fine-tuned using the target language. Therefore, only auxiliary languages are used to generate support and query sets.
An episode of the first scenario is shown in Table~\ref{table:episode_1}.

In the second scenario, in addition to auxiliary languages, we also used the target language for training. So, the support set is constructed from auxiliary language data and the query set is generated from both auxiliary and Persian language data.
Table~\ref{table:episode_2} shows an episode of the second scenario.

\begin{table*}[!ht]
	\centering
	\begin{tabular}{lr}
		\hline
		Example & Category \\
		\hline
		\hline
		\multicolumn{2}{c}{Support set (or Query set)}\\
		\hline
		\specialcellleft{{\textit{In the midst of this amazing amalgam of cultures is a passion for continuity}}\\{$\Rightarrow$ \textit{A passion for continuity is not the most important of these cultures}}}                                      & neutral  \\

		\specialcellleft{{\textit{The river plays a central role in all visits to Paris.}}\\{$\Rightarrow$ \textit{The river is central to all vacations to Paris}}}            & entailment \\
        
        \specialcellleft{{\textit{For the moment, he sought refuge in retreat, and left the room precipitately.}}\\{$\Rightarrow$ \textit{He stayed put and sat on the floor.}}}
		                                     & contradiction\\
		\hline
	\end{tabular}
	\caption{\label{table:episode_1} Example for a 3-way 1-shot episode in the first scenario. In this example we select support set and query set samples from English dataset. As support and query sets are generated similarly, only one set is shown in this table.}
\end{table*}

\begin{table*}[!ht]
	\centering
	\begin{tabular}{lr}
		\hline
		Example & Category \\
		\hline
		\hline
		\multicolumn{2}{c}{Support set}\\
		\hline
		\specialcellleft{{\textit{Recuerda que una vez mencionó que su padre era médico?}}\\{$\Rightarrow$ \textit{Ella mencionó que su padre era médico hace mucho tiempo}}}      & neutral \\
		\specialcellleft{{\textit{Dies ist etwas anderes als eine Cantina-Leuchte}}\\{$\Rightarrow$ \textit{Dies ist sicherlich keine Cantina-Leuchte}}}			 & entailment \\
		\specialcellleft{{\textit{Ensuite, il enfonce un tube respiratoire dans la gorge du patient mort.}}\\{$\Rightarrow$ \textit{Le patient vit toujours.}}}             & contradiction \\
		
		\\
		\hdashline
		English Translation & \\
		\hdashline
		\specialcellleft{\footnotesize You remember her once mentioning that her father was a doctor? \\{$\Rightarrow$ \footnotesize She mentioned her father being a doctor a long time ago.}}      & neutral \\
		\specialcellleft{{\footnotesize This is something other than a cantina fixture.}\\{$\Rightarrow$ \footnotesize{This is certainly not a cantina fixture.}}}			 & entailment \\
		\specialcellleft{{\footnotesize{Next he shoves a breathing tube down the dead patient 's throat .}}\\{$\Rightarrow$ \footnotesize{The patient is still alive.}}}             & contradiction \\
		
		\hline
		\multicolumn{2}{c}{Query set}\\
		\hline
		\specialcellleft{{\textit{Une pièce qualifie Frank Lloyd Wright de terrible ingénieur.}}\\{$\Rightarrow$ \textit{Piece a également déclaré que Wright était un bien meilleur concepteur.}}}    	& neutral \\
		\specialcellleft{{\textit{Sus rápidos oídos captaron el sonido del tren que se acercaba.} }\\{$\Rightarrow$\textit{Escuchó que el tren se acercaba rápidamente.} }}                & entailment \\
		\specialcellleft{{\footnotesize \FR{از قرن دوازدهم به بعد ارقام عربی برای نخستین بار در ایتالیا کاربرد یافت.}}\\{$\Rightarrow$\footnotesize \FR{فرانسه اولین کشوری بود که از ارقام عربی استفاده کرد.}}}               	& contradiction \\
		
		\\
		\hdashline
		English Translation & \\
		\hdashline
		\specialcellleft{{\footnotesize A piece calls Frank Lloyd Wright an awful engineer.}\\{$\Rightarrow$\footnotesize Piece also stated Wright was a much better designer.}}             & neutral \\
		\specialcellleft{{\footnotesize Her quick ears caught the sound of the approaching train.}\\{$\Rightarrow$\footnotesize She heard the train approaching fast.}}                & entailment \\
		\specialcellleft{{\footnotesize From the twelfth century onwards, Arabic numerals were first used in Italy.}\\{$\Rightarrow$\footnotesize France was the first country to use Arabic numerals.}} 	& contradiction \\
		\hline
	\end{tabular}
	\caption{\label{table:episode_2} Example for a 3-way 1-shot episode in the second scenario. In this example, the support set samples are selected from French, Spanish, and German datasets, respectively, and the query set samples are selected from French, Spanish, and Persian datasets, respectively.}
\end{table*}

\subsection{Confusion Matrices}
\label{app:confusion_matrices}

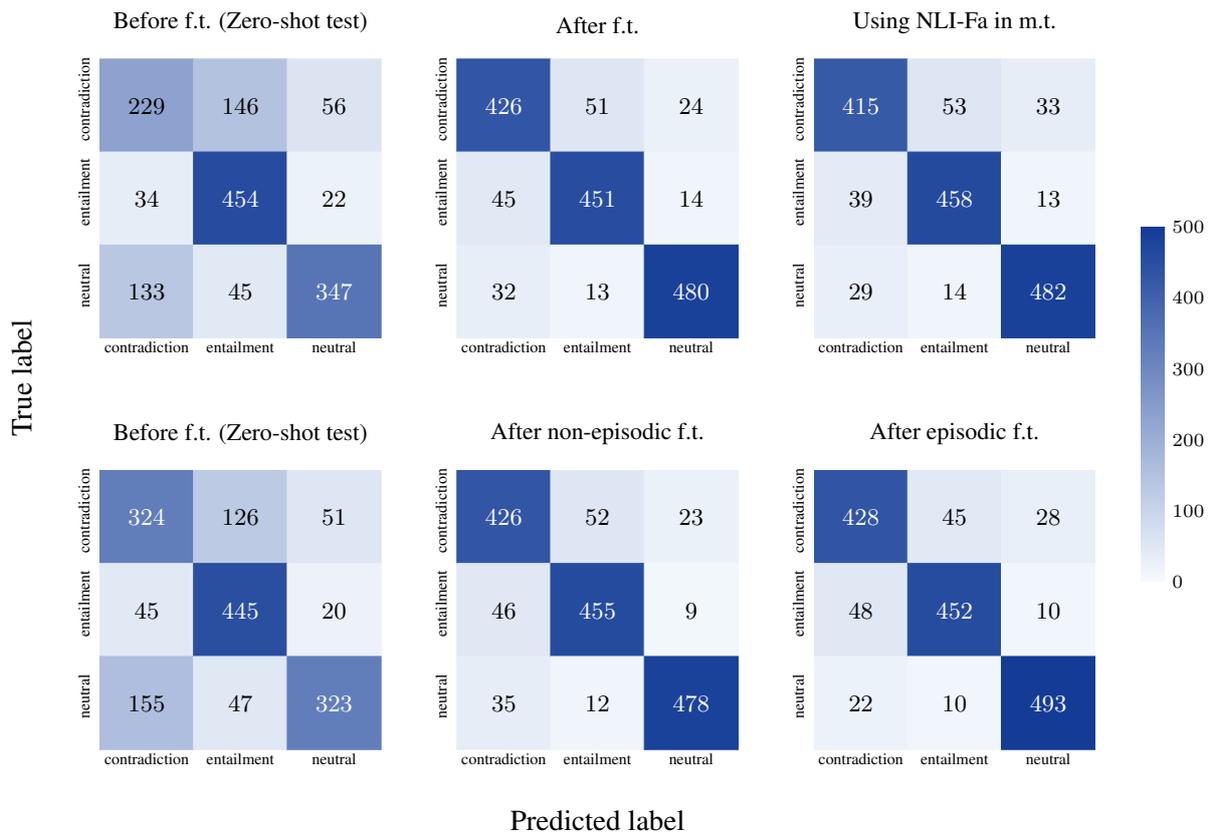
\begin{figure*}[!ht]
	\centering
    \begin{tikzpicture}
    \tikzstyle{every node}=[font=\small]
    \begin{groupplot}[
        group style = {
            group size = 3 by 2,
            vertical sep = 50pt,
            group name=scatter_combined,
        },
        width=0.33*\textwidth,
        height=0.33*\textwidth,
        colormap={bluewhite}{rgb255=(246, 250, 254) rgb255=(18, 59, 150)},
        axis line style={draw=none},
        xticklabels={contradiction, entailment, neutral},
        xtick={0,...,2},
        xtick style={draw=none},
        yticklabels={contradiction, entailment, neutral},
        ytick={0,...,2},
        ytick style={draw=none},
        every x tick label/.append style={
            font=\tiny, yshift=0.5ex
        },
        every y tick label/.append style={
            font=\tiny, yshift=0.5ex
        },
        enlargelimits=false,
        colorbar,
        yticklabel style={
          rotate=90
        },
        point meta min=0,
        point meta max=500,
        nodes near coords={\pgfmathprintnumber\pgfplotspointmeta},
        nodes near coords style={
            yshift=-7pt
        },
        colorbar to name=sharedcolorbar,
        every colorbar/.append style={
            height= \pgfkeysvalueof{/pgfplots/parent axis height} + \pgfkeysvalueof{/pgfplots/group/vertical sep},
            ytick={0,100,...,500},
            ytick style={draw=none},
            yticklabel style={font=\scriptsize},
            width=0.3cm,
            axis line style={draw=none},
        },
        visualization depends on={%
        ifthenelse(\thisrow{C}<250,100,0) \as \myv},
        nodes near coords,
        scatter/@pre marker code/.append style={/tikz/text=black!\myv},
    ]
        \nextgroupplot[
            title=Before f.t. (Zero-shot test),
            ylabel=True label,
            ylabel style={
                font=\normalsize,
                xshift=-0.45*0.33*\textwidth,
                yshift=10pt
            }
        ]
        \addplot[
            matrix plot,
            point meta=explicit,
            draw=none,
        ] table [meta=C] {
            x y C
            0 0 229
            1 0 146
            2 0 56
            
            0 1 34
            1 1 454
            2 1 22
            
            0 2 133
            1 2 45
            2 2 347
            
        };
        
        \nextgroupplot[
            title=After f.t.,
        ]
        \addplot[
            matrix plot,
            point meta=explicit,
            draw=none
        ] table [meta=C] {
            x y C
            0 0 426
            1 0 51
            2 0 24
            
            0 1 45
            1 1 451
            2 1 14
            
            0 2 32
            1 2 13
            2 2 480
            
        };
        
        \nextgroupplot[
            title=Using NLI-Fa in m.t.,
        ]
        \addplot[
            matrix plot,
            point meta=explicit,
            draw=none
        ] table [meta=C] {
            x y C
            0 0 415
            1 0 53
            2 0 33
            
            0 1 39
            1 1 458
            2 1 13
            
            0 2 29
            1 2 14
            2 2 482
            
        };
        
        \nextgroupplot[
            title=Before f.t. (Zero-shot test)
        ]
        \addplot[
            matrix plot,
            point meta=explicit,
            draw=none
        ] table [meta=C] {
            x y C
            0 0 324
            1 0 126
            2 0 51
            
            0 1 45
            1 1 445
            2 1 20
            
            0 2 155
            1 2 47
            2 2 323
            
        };
        
        \nextgroupplot[
            title=After non-episodic f.t.,
            xlabel=Predicted label,
            xlabel style={
                font=\normalsize,
                yshift=-10pt
            },
        ]
        \addplot[
            matrix plot,
            point meta=explicit,
            draw=none
        ] table [meta=C] {
            x y C
            0 0 426
            1 0 52
            2 0 23
            
            0 1 46
            1 1 455
            2 1 9
            
            0 2 35
            1 2 12
            2 2 478
            
        };
        
        \nextgroupplot[
            title=After episodic f.t.,
        ]
        \addplot[
            matrix plot,
            point meta=explicit,
            draw=none
        ] table [meta=C] {
            x y C
            0 0 428
            1 0 45
            2 0 28
            
            0 1 48
            1 1 452
            2 1 10
            
            0 2 22
            1 2 10
            2 2 493
            
        };
        \end{groupplot}
        
        \node (fig_legend) at ($(scatter_combined c1r1.center)!0.5!(scatter_combined c1r2.center)+(0.77*\textwidth,0)$){\ref*{sharedcolorbar}};
    \end{tikzpicture}
    \caption{Confusion matrices of the best-obtained model (8-shot with four auxiliary languages) in both meta-learning algorithms on the FarsTail test set. (Top): Reptile algorithm results. (Bottom): Prototypical algorithm results.}
	\label{fig:confusion}
\end{figure*}

The confusion matrices for the top-performing models (8-shot with four auxiliary languages) is depicted in Fig.~\ref{fig:confusion} showing the success of this method in improving the accuracy in all classes specially the neutral class.

\end{document}